\pdfoutput=1

\documentclass[11pt]{article}

\usepackage[final]{EMNLP2022}

\usepackage{times}
\usepackage{latexsym}

\usepackage[T1]{fontenc}

\usepackage[utf8]{inputenc}

\usepackage{microtype}

\usepackage{inconsolata}

\usepackage{booktabs}
\usepackage{graphicx}
\usepackage{hyperref}
\usepackage{float}
\usepackage{mathtools}
\usepackage{dsfont}
\usepackage{comment}

\title{Transformer Language Models without Positional Encodings\\Still Learn Positional Information}

\author{Adi Haviv$^{\tau}$ \quad Ori Ram$^{\tau}$ \quad Ofir Press$^{\omega}$ \quad Peter Izsak$^{\iota}$ \quad Omer Levy$^{\tau\mu}$ \\
\\
$^{\tau}$Tel Aviv University \hspace{0.4cm} 
$^{\omega}$University of Washington \hspace{0.6cm}
$^{\iota}$Intel Labs \hspace{0.6cm} 
$^{\mu}$Meta AI \\
\small{\texttt{ \{adi.haviv, ori.ram, levyomer\}@cs.tau.ac.il}},    \small{\texttt{ofirp@cs.washington.edu, peter.izsak@intel.com}}}

\begin{document}
\maketitle

\begin{abstract}
Causal transformer language models (LMs), such as GPT-3, typically require some form of positional encoding, such as positional embeddings. 
However, we show that LMs without any explicit positional encoding are still competitive with standard models, and that this phenomenon is robust across different datasets, model sizes, and sequence lengths.
Probing experiments reveal that such models acquire an implicit notion of absolute positions throughout the network, effectively compensating for the missing information.
We conjecture that causal attention enables the model to infer the number of predecessors that each token can attend to, thereby approximating its absolute position.
Our findings indicate that causal LMs might derive positional awareness not only from the explicit positioning mechanism, but also from the effects of the causal mask. 
\end{abstract}

\section{Introduction}
\label{sec:introduction}

The attention mechanism \cite{Bahdanau2015NeuralMT} of the transformer \cite{Vaswani2017AttentionIA} is agnostic to the position and order of tokens in the input sequence.
It is therefore common practice to inject positional information via absolute positional embeddings \cite{ Vaswani2017AttentionIA,Radford2018ImprovingLU} or relative bias factors \cite{shaw-etal-2018-self,raffel_t5, Press2021TrainST}.
Here, we demonstrate that transformer language models \textit{without} any explicit positional information can and do learn an implicit notion of absolute positions that is sufficient to achieve competitive performance.

\begin{figure}[t!]
\centering
\includegraphics[width=1\columnwidth]{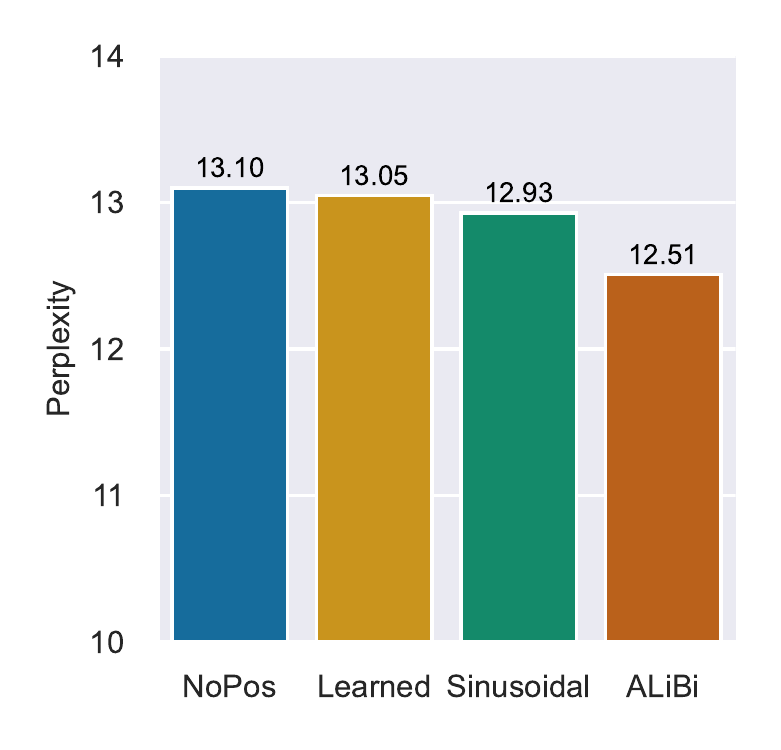}
\caption{Transformer language models trained without explicitly encoding positional information (\textit{NoPos}) approach the performance of models trained with various positional encoding methods. All models have 1.3B parameters, and are trained on an excerpt of the Pile.}
\label{fig:main_exp}
\end{figure}

We compare the performance of language models trained with no explicit positional information (\textit{NoPos} language models) to those trained with three different position-aware mechanisms, namely: sinusoidal embeddings \cite{Vaswani2017AttentionIA}, learned embeddings \cite{Gehring2017ConvolutionalST}, and ALiBi \cite{Press2021TrainST}.
Results show that NoPos models are competitive with position-aware models consistently across datasets, model sizes, and input sequence lengths (e.g., Figure~\ref{fig:main_exp}).
 
To shed light on our findings, we probe into the  position-awareness of NoPos language models, compared to models that use \textit{relative} or \textit{absolute} position mechanisms.
Specifically, we train classifiers to predict the position of a token given its representation across different layers in the network.
Our probes reveal that the NoPos model achieves similar mean absolute distance between the predicted and the expected positions, as a model with learned absolute position embeddings.

We hypothesize that this surprising behavior is tied to the \textit{causal} attention mask, which implicitly injects positional information into the self-attention layer in order to preserve the autoregressive nature of language models.
Intuitively, a model that is able to count  the predecessors of a given token can essentially infer its absolute position.
To test our hypothesis, we run similar experiments for masked language models (MLM) \cite{devlin-etal-2019-bert}, which use order-invariant attention (since no causal mask is applied).
Indeed, bidirectional models fail to converge when position information is absent, substantiating our hypothesis. To conclude, our main contributions are:
\begin{itemize}
    \vspace{-3pt}
    \item We demonstrate the robustness of the NoPos model (compared to position-aware models) with respect to model size, dataset and sequence length.
    
    
    
    \vspace{-3pt}
    \item We provide an analysis of the trained NoPos model, and show that it encoded absolute positions.
    
    \vspace{-3pt}
    \item We show that the success of NoPos models is unique to \textit{causal} language models.
    
\end{itemize}

\section{Positional Encodings}
\label{sec:methods}

Transformer models consist of interleaved self-attention and feed-forward layers, which are both order-invariant.
Therefore, to convey the order of the input tokens, some form of positional information is explicitly introduced into the model. 
\textit{Absolute positions} are commonly encoded as vectors (one for each position), which are then added to the input tokens' embeddings and fed to the first layer of the transformer.
\textit{Relative positions} are typically encoded as biases (added to attention scores) within the self-attention layers.
In this work, we consider three popular methods as baselines:

\paragraph{Learned.}
Embeddings trained to represent absolute positions \cite{memnets, Gehring2017ConvolutionalST}.
Learned positional embeddings are commonly used in MLMs \cite{devlin-etal-2019-bert, roberta} as well as in large autoregressive language models, such as GPT-3 \cite{gpt3}. 

\paragraph{Sinusoidal.}
Constant vectors computed by a non-parametric function of the input token's absolute position.
Sine and cosine functions of different frequencies are used, such that each dimension of the positional encoding corresponds to a sinusoid.
Sinusoidal embeddings were introduced in \citet{Vaswani2017AttentionIA} for machine translation, and are also used in language modeling  \cite{DBLP:conf/iclr/BaevskiA19}.

\paragraph{ALiBi.}
Attention with LInear BIases \cite{Press2021TrainST} injects information about the relative distances between tokens by adding negative biases to attention scores, which grow linearly with the distance between each pair of tokens.

\section{Experiment Setup}
\label{sec:experiments}

Intuitively, encoding positional information explicitly is crucial for enabling transformer language models to predict the next token in a sequence.
To test this intuition, we compared the validation set perplexity of models trained from scratch with no explicit positional information (denoted as \textit{NoPos}) to those trained with the various positional encoding methods discussed in Section~\ref{sec:methods}.
We investigated the canonical WikiText-103 setting \cite{MerityX0S17, DBLP:conf/iclr/BaevskiA19}, as well as a newer, large-scale setting based on the Pile corpus \cite{pile} on model architectures inspired by \citet{gpt3}, where we cover a spectrum of models sizes and sequence lengths.


\paragraph{The Canonical Setting (WikiText-103).}
The WikiText-103 corpus \cite{MerityX0S17} consists of over 100 million words extracted from a set of high-quality Wikipedia articles.
The corpus is tokenized at the word level, resulting in a vocabulary of over 267K tokens. For this corpus, we used the adaptive embedding transformer model of \citet{DBLP:conf/iclr/BaevskiA19}, which contains 16 transformer layers with 1024 model dimensions, 4096 feed-forward dimensions, and 8 attention heads. Overall, this model has 247M parameters in total.
We trained with their exact optimization hyperparameters, as implemented in \texttt{fairseq} \cite{ott-etal-2019-fairseq}, with the exception of the input sequence length, which was shortened to 512 tokens (instead of 3072), as in \citet{Press2021TrainST}.
See App.~\ref{apx:hyperparameters} for detailed hyperparameters.

\paragraph{The Large-Scale Setting (The Pile).}
\label{pile}
The Pile \cite{pile} is an 800GB English text dataset composed of Common Crawl and 22 other diverse sources. For our experiments, we used 2 out of 30 shards;\footnote{Shards 00 and 01 can be downloaded from: \url{https://the-eye.eu/public/AI/pile/train/}} of these, we filtered out the GitHub and DM Mathematics sources and removed the shortest 1\% and longest 1\% of examples from each source to reduce noise. We used GPT-2's tokenizer \cite{gpt2} to convert the text into token sequences over a vocabulary of 50K tokens.
We randomly sampled a validation set of 2000 documents (2.6M tokens) from the corpus, while the remaining 15M documents (21B tokens) comprised the training set. The baseline model in this setting follows the 1.3B parameter architecture of \citet{gpt3}, also known as GPT-3 XL: 24 transformer layers with 2048 model dimensions, 8192 feed-forward dimensions, and 32 attention heads.
The default input sequence length is 1024 tokens. We refer to App.\ref{apx:hyperparameters} for detailed hyperparameters. 

To demonstrate the consistency of our results in different settings, we perform two scaling experiments. We first scale the model size by experimenting with the small (125M parameters), medium (350M parameters), large (760M parameters) and the XL (1.3B parameters) variants of the \citet{gpt3} architecture on the Pile settings. In addition, we evaluate the effect of varying the sequence length using the XL (1.3B parameter) model. Specifically, we experiment with sequences of lengths $\{256,512,1024,2048\}$.

Last, to shed additional light on differences between the NoPos model to other methods, we compare the model's performance on different parts of the sequence. Details of this analysis and results are given in App.~\ref{apx:seq_split}.

\section{Results}
\label{sec:results}

\begin{table}[t]
    \small
    \centering
    \begin{tabular}{@{}lrr@{}}
    \toprule
    & \textbf{WikiText-103} & \textbf{The Pile} \\
    \midrule
    NoPos & 20.97 & 13.10 \\
    Learned & 20.42 & 13.05 \\
    Sinusoidal\qquad\qquad & 20.16 & 12.93 \\
    ALiBi & 19.71 & 12.51 \\
    \bottomrule
    \end{tabular}
    \caption{Validation set perplexity of transformer language models trained with various positional encoding methods. The WikiText-103 setting \cite{MerityX0S17} uses the model of \citet{DBLP:conf/iclr/BaevskiA19} on sequences of 512 tokens, while the Pile settings \cite{pile} uses a more recent 1.3B parameter architecture \cite{gpt3} over 1024 token sequences.}
    \label{tab:results}
\end{table}

Table~\ref{tab:results} compares the performance of training LMs with different position encoding methods.
We observe that NoPos LMs approach the performance of the other models, with gaps of 0.55 (WikiText-103) and 0.05 (the Pile) perplexity from models with \textit{learned} positional embeddings.
In the Pile setting, performance differences between \textit{NoPos}, \textit{Learned}, and \textit{Sinusoidal} are small both in absolute terms and with respect to their difference with \textit{ALiBi}.
In the WikiText-103 setting, performance gaps are wider but still modest with respect to random seed variance.\footnote{For context, \citet{press-etal-2020-improving} report that training the sinusoidal model with inputs of length 3072 on WikiText-103 with 5 different seeds can result in gaps of up to 0.9 perplexity between runs (0.34 standard deviation).}
These results strongly suggest that training transformer language models without explicit positional encoding is indeed possible. 

Table~\ref{tab:size} explores the effects of scaling the number of parameters in the Pile setting.
While smaller models benefit from fixed, non-parametric positional encodings (\textit{Sinusoidal} and \textit{ALiBi}), these performance gaps narrow in larger models.
Table~\ref{tab:length} shows the effect of varying the sequence length in the same setting.
In this experiment, the gaps between \textit{NoPos}, \textit{Learned}, and \textit{Sinusoidal} remain almost constant, while the benefit of using \textit{ALiBi} increases as sequences become longer.
Overall, we show that transformer language modeling without explicit positional encoding is robust to the selection of corpus, model size, and sequence length.

As training models at the 1.3B parameter scale is resource-intensive, we publicly release our trained models for future research and analysis.\footnote{\url{https://github.com/adihaviv/NoPos}}

\begin{table}[h]
    \small
    \centering
    \begin{tabular}{@{}lrrrr@{}}
    \toprule
    \textbf{Model Size}\qquad\qquad & \textbf{125M} & \textbf{350M} & \textbf{760M} & \textbf{1.3B} \\
    \midrule
    NoPos & 22.15 & 16.87 & 14.29 & 13.10 \\
    Learned & 22.04 & 16.84 & 14.21 & 13.05 \\
    Sinusoidal & 21.49 & 16.58 & 14.04 & 12.93  \\
    ALiBi & 19.94 & 15.66 & 13.53 & 12.51 \\
    \bottomrule
    \end{tabular}
    \caption{Validation set perplexity on the Pile, as a function of positional encoding method and model size. All models operate on sequences of 1024 tokens. Smaller models benefit from fixed, non-parametric positional encodings (\textit{Sinusoidal} and \textit{ALiBi}), but these performance gaps diminish as the models scale up. }
    \label{tab:size}
\end{table}

\begin{table}[H]
    \small
    \centering
    \begin{tabular}{@{}lrrrr@{}}
    \toprule
    \textbf{Seq Length} & \textbf{256} & \textbf{512} & \textbf{1024} & \textbf{2048} \\
    \midrule
    NoPos & 14.98 & 13.82 & 13.10 & 12.87 \\
    Learned & 14.94 & 13.77 & 13.05 & 12.72 \\
    Sinusoidal & 14.84 & 13.66 & 12.93 & 12.62 \\
    ALiBi & 14.65 & 13.37 & 12.51 & 12.06 \\
    \bottomrule
    \end{tabular}
    \caption{Validation set perplexity on the Pile, as a function of positional encoding method and sequence length. All models have 1.3B parameters.
    The performance differences between \textit{NoPos}, \textit{Learned}, and \textit{Sinusoidal} are consistently small, while \textit{ALiBi} slowly becomes more beneficial as sequences become longer.}
    \label{tab:length}
\end{table}

In a Concurrent work, \citet{anonymous2022what} makes a similar observation in one of their ablation experiments and further show that NoPos models gain competitive performances for downstream tasks as well. Specifically, they evaluated 27 diverse downstream tasks. They showed that the NoPos model reached an average accuracy of $41.23\%$ over all tasks, comparing to \textit{Learned} and \textit{ALiBi} who gained $41.72\%$ and $43.70\%$ respectively.

\section{Analysis}
\label{sec:analysis}
In this section, we examine whether the NoPos model is able to encode positional information and show that such information is essential for its success.

\begin{figure}[t]
    \centering
    \includegraphics[width=\columnwidth]{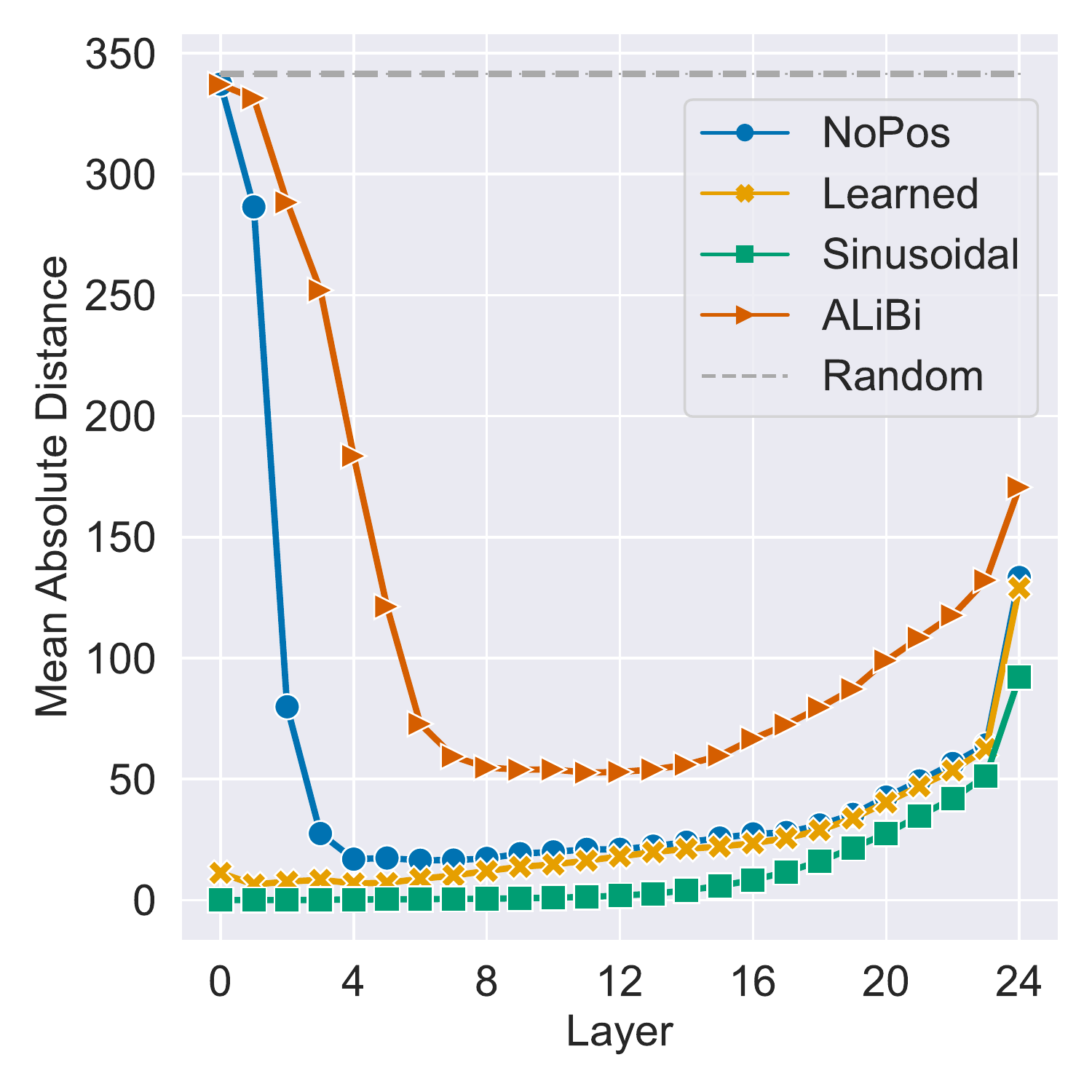}
    \caption{Through probing, we find that the NoPos model behaves similarly to models that use absolute learned position embeddings. We evaluated performance using mean absolute distance on 1.3B parameter models trained on the Pile.}
    \label{fig:probe}
\end{figure}

\paragraph{NoPos models acquire positional information} Do NoPos LMs learn some form of positional encoding to compensate for the absence of explicit positional modeling?
To answer this question, we probe each layer of our trained models\footnote{We used the 1.3B parameter models trained over 1024-token sequences of the Pile (Section~\ref{sec:experiments}).} for positional information.
Specifically, we use the tokens' last hidden representation after each transformer layer, produced by the evaluated LM, and train a 2-layer feed-forward ReLU network to predict the absolute position (0 to 1023) of each token (i.e., as a multiclass classification problem). Notably, we do not change the weights of the evaluated LMs and thus, do not provide any position information of the tokens to the LM in this experiment, which ensures the validity of our findings. 

Each layer's probe was trained separately (hyperparameters are provided in App.~\ref{apx:hyperparameters}).
As a soft accuracy metric, we measured the mean absolute distance between the probe's prediction and the token's actual position.

Figure~\ref{fig:probe} shows that even though NoPos model starts, as expected, with no positional information in the first layer (on par with a random baseline), it becomes position-aware within four layers and appears to contain more positional information than ALiBi. By the middle layer, NoPos can predict absolute positions about as well as the model with learned positional embeddings. 
Finally, we observe that all models shed off a significant amount of positional information in the final layers, in line with the findings of \citet{voita-etal-2019-bottom}.
Overall, the probe reveals that the NoPos models learn an implicit notion of absolute positions. 

\begin{figure}[t]
\centering
\includegraphics[width=1\columnwidth]{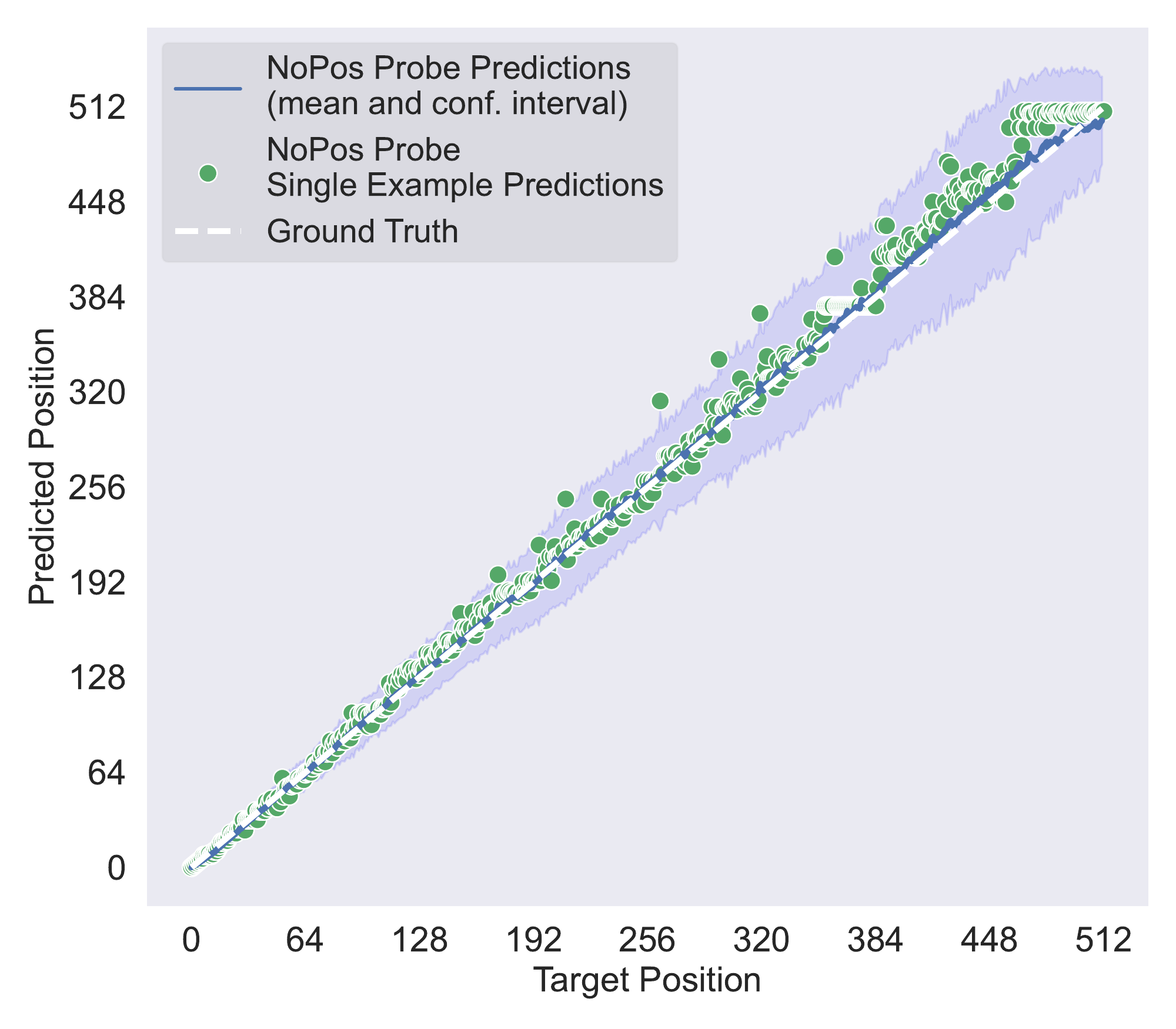}
\caption{A visualization of the absolute position predictions of a probe trained over a NoPos language model. The blue line shows the mean of the generated predictions for every target position and the blue area represents the 95\%-confidence interval. The predictions for a single random sequence are depicted as green dots.}
\label{fig:100_examples}
\end{figure}

To elucidate what positional information the NoPos model learns, we visualize the predictions of the probe. We examine a sample of 100 predictions from the validation set of the best-performing probe trained over the NoPos model. Figure~\ref{fig:100_examples} shows the predictions over the 512 token sequences sampled randomly from the validation set and a single example from the same set. We observe that the probe is more accurate at the beginning of the sequence, but becomes fuzzier as it progresses.

\paragraph{Positional information matters}\label{par:analysis_pos_matters} 
NoPos is able to infer absolute positions, but are they necessary? We answer this using a trained NoPos model. Instead of computing the loss over the entire sequence, we select a single random token, shuffle the previous tokens that it is conditioned on, and compare to a baseline where the prefix remains intact. 
We find that in the case where the suffix is shuffled, the average token-level loss increases dramatically (from $\sim$4 to $\sim$11).
Details of this experiment are given in App.~\ref{apx:word_order}.

This finding indicates that the NoPos model indeed uses the positional information it acquires, as otherwise we would expect similar loss values in these two settings.

\section{Conjecture}
\label{sec:hypothesis}

How do transformers without explicit positional encoding learn absolute positions?
We conjecture that the \textit{causal attention} in autoregressive  transformer language models allows them to predict the number of attendable tokens at each position, i.e. the number of tokens in the sequence that precede the current one.
Such a mechanism could effectively encode the \textit{absolute} position of each token into its vector representation.
Indeed, our analysis (Section~\ref{sec:analysis}) reveals that some notion of absolute positions exists in the hidden layers of language models even when they are trained without explicit positional encoding, and that this information is acquired throughout the first few layers.
On the other hand, bidirectional transformer encoders (which are used in masked language modeling, e.g. \citealt{devlin-etal-2019-bert}) do not contain causal attention masks or any other limitation on the attention mechanism; thus, they should be unable to learn absolute positions without explicit positional encoding.
We tested this corollary by training a masked language model based on RoBERTa large \cite{roberta} on the Pile (see App.~\ref{apx:hyperparameters} for hyperparameters). Table~\ref{tab:mlm} shows that, indeed, the NoPos model has significantly worse perplexities than the position-informed baselines.
This result echoes the findings of \citet{sinha-etal-2021-masked}, who also observed that MLMs without positional embeddings suffer significant performance degradation.

\begin{table}[t]
    \small
    \centering
    \begin{tabular}{@{}lr@{}}
    \toprule
    & \textbf{MLM Perplexity} \\
    \midrule
    NoPos & 147.18 \\
    Learned & 4.06 \\
    Sinusoidal\qquad\qquad & 4.07  \\
    ALiBi & 4.00 \\
    \bottomrule
    \end{tabular}
    \caption{Validation set perplexity of \textit{masked} language models \cite{devlin-etal-2019-bert} trained with various positional encoding methods on an excerpt of the Pile \cite{pile}. The model architecture is based on RoBERTa large \cite{roberta}, and processes 128 tokens per sequence.
    While position-aware models converge to very low perplexities, training without positional encodings (\textit{NoPos}) fails.}
    \label{tab:mlm}
\end{table}

\section{Related Work}
\label{sec:related_work}
While there has been ample research on positional encoding variants, there has been relatively little prior work that investigate  models' ability to infer positions implicitly. Prior to our work, \citet{Irie2019LanguageMW} explored transformer language models for speech recognition and found that such models, when trained without positional encoding, outperform those trained with sinusoidal embeddings. In addition, a focused language modeling experiment by Stella Rose Biderman\footnote{\url{https://twitter.com/BlancheMinerva/status/1394089508723900422}} showed that the NoPos method attains similar results to other position embedding methods; however, that experiment was on a small 350M parameter model trained on a small character-level dataset (enwik8). Here we show that this result holds across multiple datasets and model sizes, provide an analysis of the model's internal representations, and hypothesize how this phenomenon could occur.
\section{Conclusion}
\label{sec:conclusion}
We show that, contrary to popular belief, transformers language models do learn positional information even when are not provided with any explicit positional encoding.
Our experiments systematically demonstrate that this phenomenon is robust across different language modeling settings, and that one can approximate the absolute position of each token from the model's internal representations to a surprising degree.
However, this phenomenon does not extend to transformer encoders trained on the MLM objective.
We conjecture that the causal attention mechanism, which limits attention in one direction of the sequence, is responsible for implicitly imbuing the transformer with positional information.

\section{Limitations}
Our work explores language models in the 125M to 1.3B parameter range. We show that as parameter count increases the gap between the NoPos method and the other position methods narrows. This trend leads us to believe that our findings should hold for even larger models, but the current biggest models are more than one hundred times bigger (in terms of parameters) than our 1.3B parameter models, and so the results in that setting can be unexpected. 
In addition, training models at the 1.3B parameter scale is resource-intensive and might hinder reproducibility. We therefore release our trained models. In Addition, when comparing the perplexity of NoPos to other models, although the margins are very small, NoPos is always slightly worse, suggesting that the inductive bias of positional encoding is indeed important.

\section*{Acknowledgements}
This work was supported by Intel Corporation, Meta Platforms Inc and Deutsch Foundation.

\bibliography{anthology,custom}
\bibliographystyle{acl_natbib}

\clearpage
\appendix

\section{NoPos Performance Across Different Segments of the Input}
\label{apx:seq_split}
To shed more light on the findings shown in section~\ref{sec:results}, we explore 
whether there are parts of the sequence that the NoPos model better predicts compared to other positional methods (e.g., is the NoPos model performs better at the beginning or the end the sequence). We compute the model's perplexity in different parts of the sequences. Specifically, we split each input sequence into eight consecutive segments and compute the perplexity for each segment separately.

\begin{figure}[H]
\centering
\includegraphics[width=1\columnwidth]{ 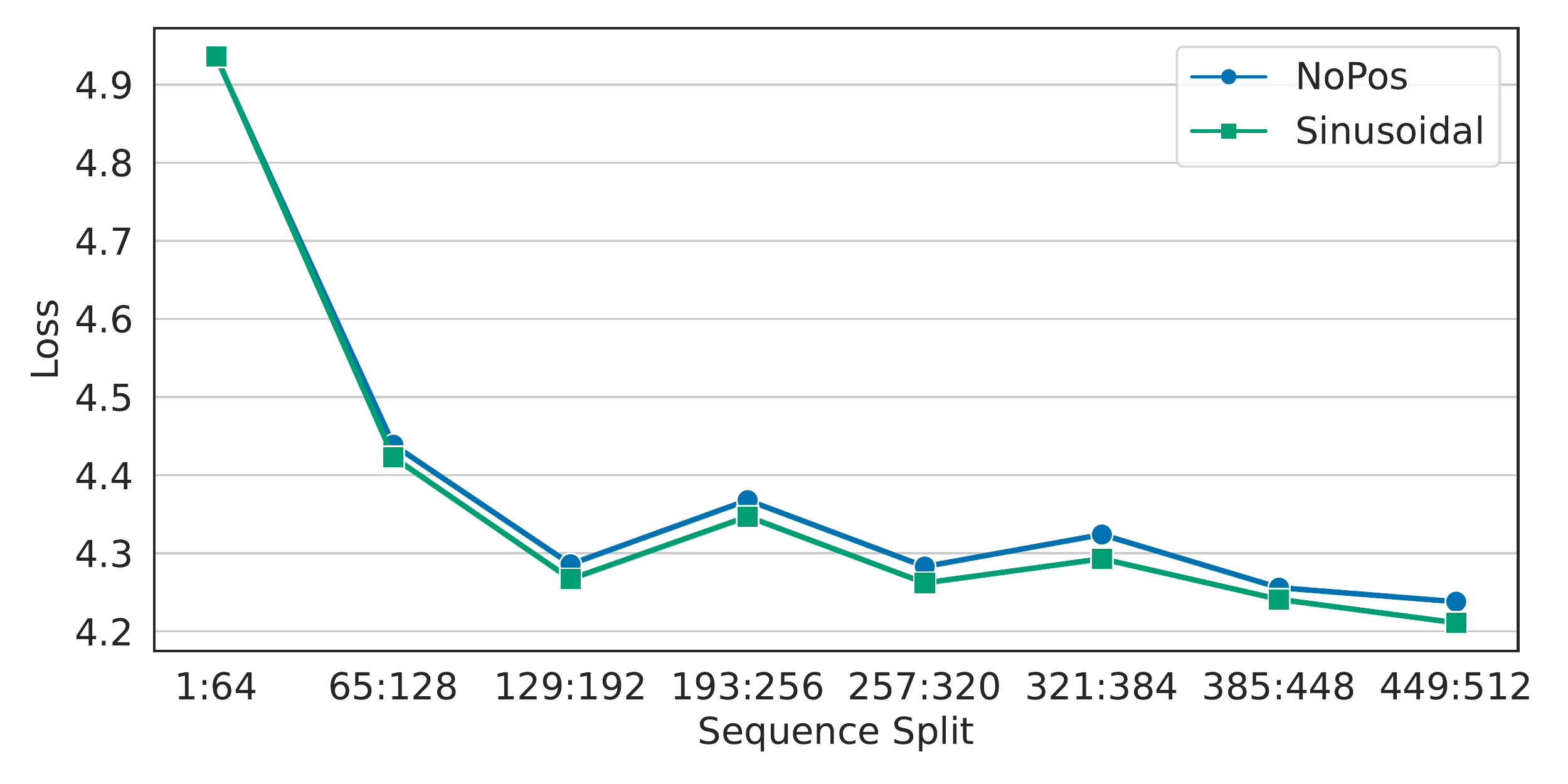}
\caption{NoPos model shows similar performances on each part of the sequence, comparing to the baseline \textit{Learned} absolute position encoding.}
\label{fig:part_ppl}
\end{figure}

We evaluate the NoPos and Learned 1.3B parameter models trained on the Pile, with input sequence length of 1024, and use the standard validation set. Figure \ref{fig:part_ppl} shows the results of this experiment. The NoPos model performs similarly or slightly worse than the baseline model on all input parts. 

\section{Word Order Analysis}
\label{apx:word_order}
Is positional information necessary for language modeling, or does the order of the input tokens not matter? To answer this, we conduct the following experiment: instead of computing the loss on the complete sequence, we pick a specific token in the sequence. The next token prediction is conditioned on the previous tokens in the sequence, and so we shuffle the order of the tokens in the prefix and compute the loss only for that specific token. We repeat the experiment with the original, un-shuffled prefix sequence as the baseline and compare the results.  

\begin{figure}[H]
\centering
\includegraphics[width=1\columnwidth]{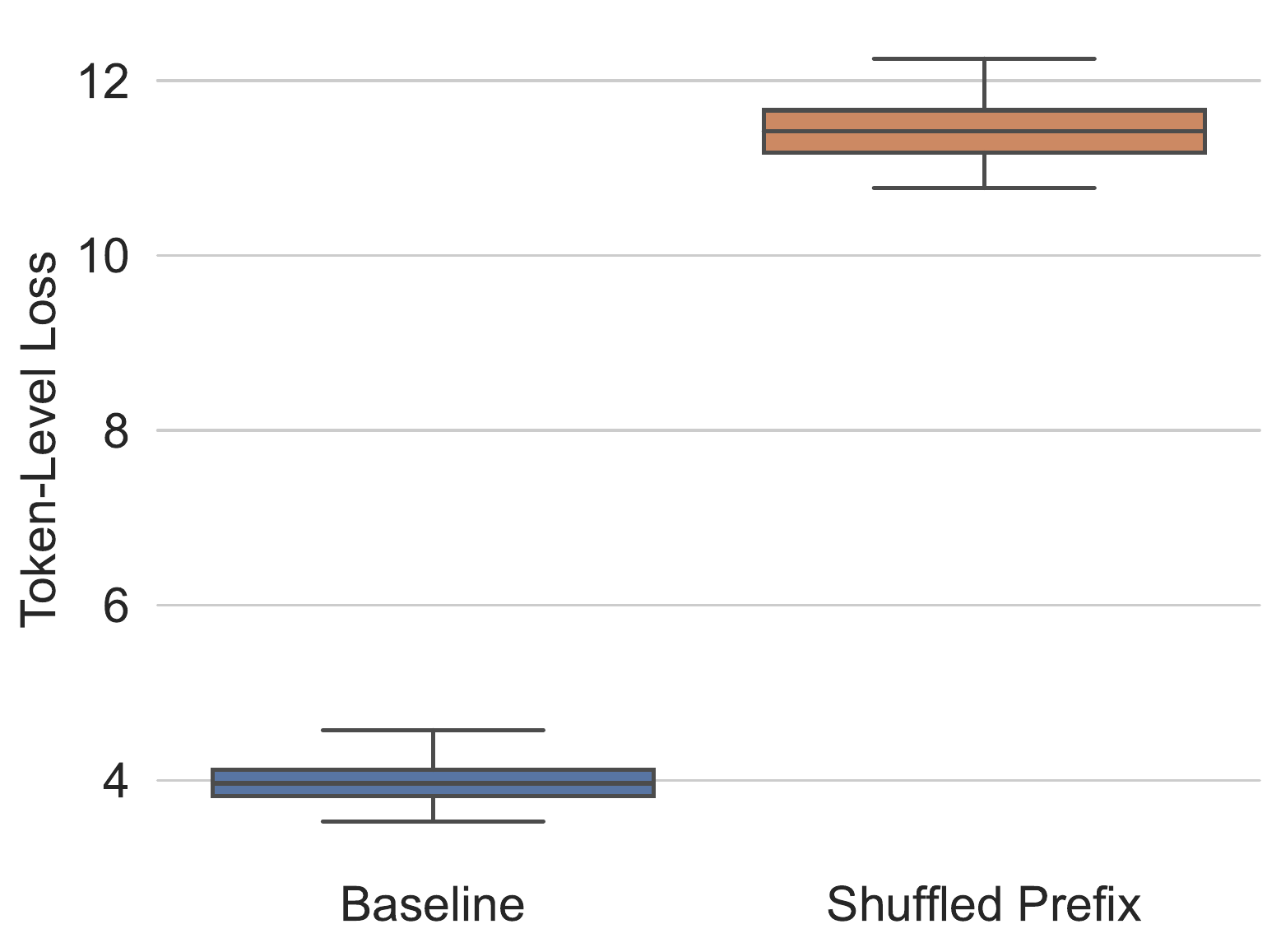}
\caption{Shuffling input tokens (for causal langauge modeling) leads to a massive degradation in token-level loss.}
\label{fig:shuffle_exp}
\end{figure}

The experiment was conducted on the NoPos model with an input sequence length of 512 using the WikiText-103 dataset. We randomly sample an index between 5 and 512 for the token we pick from each input sequence from the validation set. Figure \ref{fig:shuffle_exp} shows the results of this experiment for 100 different inputs.
These results clearly show that the transformer language model's next word predictions are not order-invariant.

\section{Hyperparameters}
\label{apx:hyperparameters}

Table~\ref{tab:hyperparameters} provides the optimization hyperparameters for each one of our experiments, and Table~\ref{tab:scale_hyperparameters}
shows the model hyperparameters in the modern (Pile) setting.

\begin{table*}[t]
    \centering
    \small
    \begin{tabular}{@{}lrrrr@{}}
    \toprule
    & \textbf{WikiText-103} & \textbf{The Pile} & \textbf{Probe} & \textbf{Masked LM}\\
    \midrule
    Sequence Length & 512 & 1024 & 1024 & 128 \\
    Optimizer & NAG & Adam & Adam & Adam \\
    Peak Learning Rate & 1 & 2e-3 & 2e-3 & 1e-3 \\
    Warmup Steps & 16,000 & 500 & 500 & 500 \\
    Total Steps & 286,000 & 10,000 & 10,000 & 10,000 \\
    Tokens per Batch & 72,000 & 256,000 & 64,000 & 1,024,000 \\
    Dropout & 0.3 & 0 & 0 & 0.1 \\
    Weight Decay & 0 & 0.01 & 0.01 & 0.01 \\
    \bottomrule
    \end{tabular}
    \caption{The optimization hyperparameters used in this work. The \textit{NAG} optimizer refers to Nesterov accelerated gradient \cite{Nesterov1983AMF}, and Adam refers to \cite{adam}.}
    \label{tab:hyperparameters}
\end{table*}

\begin{table*}[t]
    \centering
    \small
    \begin{tabular}{@{}lrrrr@{}}
    \toprule
    & \textbf{125M} & \textbf{350M} & \textbf{760M} & \textbf{1.3B}\\
    \midrule
    Layers & 12 & 24 & 24 & 24 \\
    Model Dimensions & 768 & 1024 & 1536 & 2048 \\
     Feed-forward Dimensions  & 3072 & 4096 & 6144 & 8192 \\  
    Attention Heads & 12 & 16 & 16 & 32 \\
    \bottomrule
    \end{tabular}
    \caption{The models hyperparameters by size.}
    \label{tab:scale_hyperparameters}
\end{table*}

\end{document}